\documentclass{ecai}
\usepackage{graphicx}
\usepackage{latexsym}

\usepackage{subcaption}
\usepackage{hyperref}
\usepackage{url}
\usepackage{tabularx}
\usepackage{multirow}
\usepackage{color}
\usepackage{arydshln}
\usepackage{xcolor,pifont}

\usepackage{amsmath}
\usepackage{amsthm}
\usepackage{booktabs}
\usepackage{algorithm}
\usepackage{algorithmic}
\usepackage[switch]{lineno}

\newcommand{\cf}{\emph{cf.}~}

\newcommand{\etal}{\emph{et al.}~}

\newcommand*{\imgsmall}[1]{%
    \raisebox{-.3\baselineskip}{%
        \includegraphics[
        height=\baselineskip,
        width=\baselineskip,
        keepaspectratio,
        ]{#1}%
    }%
}

\begin{document}

\begin{frontmatter}

\title{Revision Transformers: \\Instructing Language Models to Change their Values}

% \author{\snm{anonymous}}
\author[A]{\fnms{Felix}~\snm{Friedrich}\thanks{Corresponding Author. Email: friedrich@cs.tu-darmstadt.de}}
\author[A]{\fnms{Wolfgang}~\snm{Stammer}}
\author[A,B]{\fnms{Patrick}~\snm{Schramowski}} % use of \orcid{} is optional
\author[A,C]{\fnms{Kristian}~\snm{Kersting}}

\address[A]{AIML lab (TU Darmstadt) and hessian.AI}
\address[B]{DFKI and LAION}
\address[C]{DFKI and Centre for Cognitive Science (TU Darmstadt)}

\begin{abstract}
    Current transformer language models (LM) are large-scale models with billions of parameters. They have been shown to provide high performances on a variety of tasks but are also prone to shortcut learning and bias. Addressing such incorrect model behavior via parameter adjustments is very costly. This is particularly problematic for updating dynamic concepts, such as moral values, which vary culturally or interpersonally. In this work, we question the current common practice of storing all information in the model parameters and propose the Revision Transformer (RiT) to facilitate easy model updating. The specific combination of a large-scale pre-trained LM that inherently but also diffusely encodes world knowledge with a clear-structured revision engine makes it possible to update the model's knowledge with little effort and the help of user interaction. We exemplify RiT on a moral dataset and simulate user feedback demonstrating strong performance in model revision even with small data. This way, users can easily design a model regarding their preferences, paving the way for more transparent AI models.
\end{abstract}

\end{frontmatter}

\section{Introduction}
The massive amount of available data, computational resources, and research advances have recently led to the development of novel large-scale models. Showing promising SOTA results on many challenging benchmarks, some might consider these models to represent an important step towards the long-standing goal of artificial general intelligence. Regardless of whether this is true or not, there is still some work to be done. %, e.g in terms of debiasing. %Particularly concerning subjective topics, e.g.~sparser data. 
For instance, these models have been shown to be inherently affected by bias and can act as \textit{stochastic parrots}. In particular, large-scale pre-training on huge amounts of (uncurated) data, which is a common current practice, can lead to these models reflecting unwanted societal biases \cite{parrots_bender}.

The first important step towards mitigating model bias is to detect it. However, the process of large-scale pre-training makes it difficult for an individual to inspect the training data. Due to this, several recent approaches have focused on proper documentation of models and data already from the beginning of the process \cite{gebru_datasheets,mitchell_modelcards}. 
% propose to document the creation of a dataset and provide a \textit{datasheet} for the dataset used. Furthermore, \cite{mitchell_modelcards} suggest also reporting the whole model in a model card to get further insights into potential bias inherent to the model.
Yet another approach is to filter data prior to training \cite{schramowski_q16}. Unfortunately, this approach is unavailable for the bulk of end users, which in general, are missing the necessary resources for training, leading them to rely on pre-trained models. 

\begin{figure*}[t]
    \centering
    \includegraphics[width=0.85\textwidth]{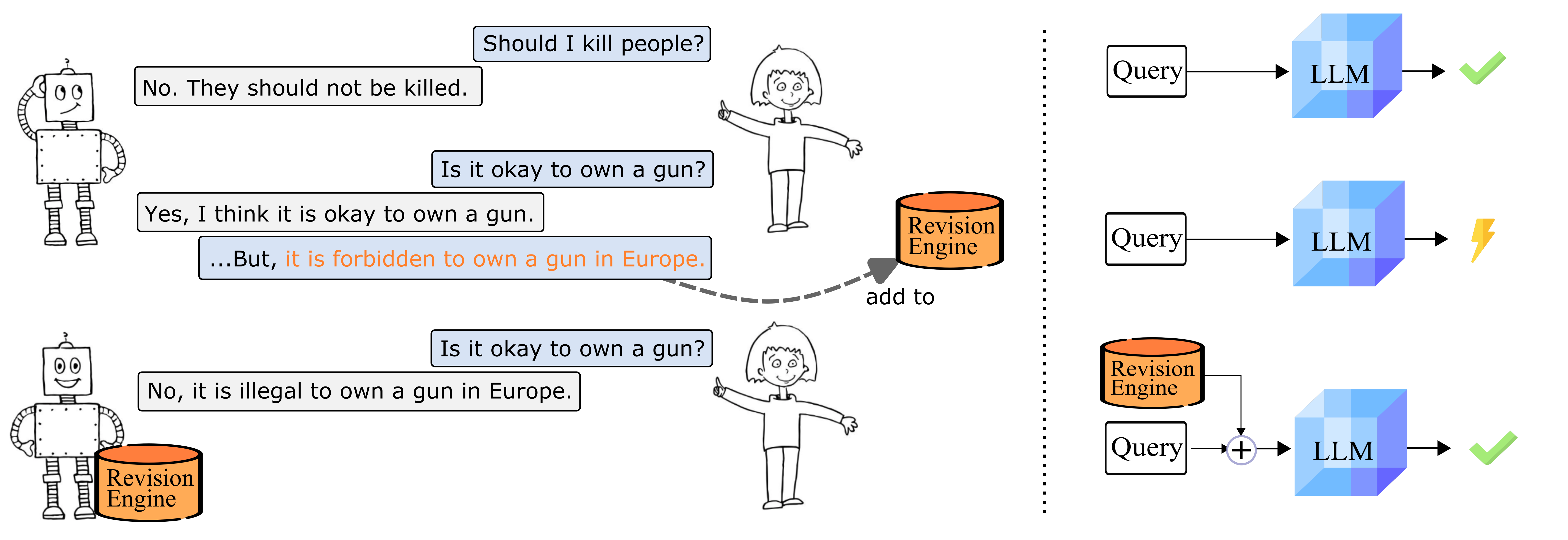}
    \caption{Human-AI conversation with user revision. A European user queries an LLM to check whether it is aligned with their values. By adding context (orange color), they revise the LLM to generate an answer regarding their cultural and personal preferences.}
    \label{fig:motivation_img}
\end{figure*}
Importantly, neither of these approaches offers a useful technique for handling subjective and oftentimes data-scarce topics, e.g.~correcting a model's moral knowledge representations.
In their recent work, \cite{delphi_jiang} retrain a large-scale LM (LLM) on moral data to show its ability to align with the moral values represented in a given dataset. The authors hereby propose an oracle-like model. This, however, has several drawbacks, most important of which has previously been mentioned: retraining a model in this way is infeasible for the majority of end users. 
Moreover, \cite{fraser-etal-2022-moral} show that although this finetuned model, Delphi, is generally aligned with the values represented in the given dataset, i.e.~the annotators' values, it remains inconsistent e.g.~in the Trolley dilemma. Furthermore, other approaches \cite{forbes-etal-2020-social} trained a model from scratch to align with user values. While this worked to a limited extent, their approach suffers from the same problem: training is costly and often unavailable.

Overall, moral values are highly subjective and vary interpersonally. Finetuning a model to incorporate the moral values represented in one dataset can likely never fully satisfy a population's diverse demands on societal and moral values (\textit{c.f.} example depicted in the left half of Fig.~\ref{fig:motivation_img}). 
% Furthermore, as other works in the field of deep learning have shown, finetuning can come at the undesired cost of catastrophic forgetting \cite{goodfellow2013_catastrophic}. %, and even worse, it is so costly that only a few tech companies can conduct it. 
Even beyond this, the temporal degradation of values is a major problem with large-scale pre-training \cite{temporal_dhingra}, making it necessary for future AI models to be able to regularly expand their knowledge in order to keep up to date with changing societal values. 
All in all, these issues suggest the necessity of %in this work we therefore call for an easy, online
revision approaches that go beyond parameter retraining. 

In this work, we, therefore, propose a novel framework, the Revision Transformer (RiT), that enables to interactively revise a model to align it with user values. In an information retrieval-based approach, RiTs extend current parametric transformer architectures with a non-parametric, interactive revision mechanism, we call \textit{revision engine}.
% , to guide the model behavior. %More precisely, RiTs extend the standard transformer architecture with a revision engine, an information retrieval approach., in which ... 
% External information helps address some vanilla transformer defects as it can be directly revised or expanded, and accessed information can be inspected and interpreted. 

Fig.~\ref{fig:motivation_img} briefly sketches some of the important properties and use cases of RiTs. If a T5 model \cite{colin_t5} is queried with ``Should I kill people?'', it provides an answer that is aligned with a user's values (\imgsmall{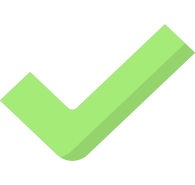}), illustrating the findings of \cite{schramowski2022nmi_moral} that LLMs already possess an initial moral dimension and a notion of right and wrong. However, the same model queried, e.g., on a more controversial topic, can also provide answers that are unaligned with a user's values (\imgsmall{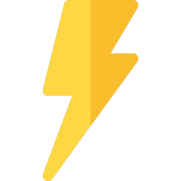}). %E.g. while some people might agree that it is ok to own a gun, others will not, depending on their surrounding culture, country's laws or simply their personal opinion. 
% The user can next provide external context to RiT's \textit{revision engine} 
% which changes the model's answer regarding their preference. This way, one 
In this case, with a RiT, a user can revise or extend the values stored in the model parameters via an external revision corpus within the \textit{revision engine}. This engine ultimately acts as an editing mechanism to store corrective knowledge provided by the user and makes it possible for RiTs to enable users to set up their individual revision engines from scratch.
% and can even let a user know if it is uncertain about an output, overall allowing to iteratively integrate the user into the revision loop. 

% Fig.~\ref{fig:motivation_img} briefly sketches some of the important properties and use cases of RiTs. If a T5 model \cite{colin_t5} is queried with ``Should I kill people?'', it answers, "No. They should not be killed" illustrating the findings of \cite{schramowski2022nmi_moral} that LLMs already possess an initial moral dimension and notion of right and wrong. However, if we query the same model with ``Is it okay to own a gun?'', it answers, "Yes, it is okay to own a gun.". While some people might agree with this, others will not, depending on their surrounding culture, country's laws or simply their personal opinion. The user can next provide external context  
% % ``It is forbidden to own a gun in Europe. Is it okay to own a gun?''
% ``It is forbidden to own a gun in Europe.'' which changes the model's answer. This way, one can revise or extend the values stored in the model parameters with an external revision corpus, the \textit{revision engine}. This engine essentially acts as a revision mechanism to store corrective knowledge provided by the user.\ws{add the first example in figure as well? shorten out parts of the chat bubbles from the main text} Our idea is thus to enable users to interactively set up their individual revision engines from scratch. A RiT can even let a user know if it is uncertain about an output, and thus allows to iteratively integrate the user into the revision loop.

An underlying question that our approach poses is: \textit{Should all information and values of a system be stored solely in model parameters through large-scale pre-training?}
It thus stands in line with other recent works that have shown the advantage of combining parametric LLMs with retrieval mechanisms \cite{rag_lewis,retro_borgeaud}.
Where previous works have focused on using large corpora of factual knowledge and benchmarked on factual QA data through end-to-end training, we shift the focus to non-factual knowledge and regimes with data sparsity. 

%To summarize, our work highlights the advantages of
% Our contributions are as follows: we (i) propose a novel framework, RiT, to interactively revise an LLM, we (ii) simulate user feedback and show strong performance, particularly in the context of small data, and overall we (iii) pave the way for new methods that enable easily adaptable transformer models.

Our contributions are as follows: we (i) propose a novel framework, RiT, to interactively revise an LLM, we (ii) show strong performance of RiT in model revision, particularly in the context of small data, and finally we (iii) leverage user feedback in an iterative fashion, further improving the RiT performance\footnote{Our code is publicly available \href{https://github.com/ml-research/Revision-Transformer}{here}.}.

% We proceed as follows. We start by briefly reviewing related work of revising LLMs. Then we introduce the Revision Transformer, including the interactive revision engine. Before concluding, we touch upon the results of our experimental evaluation and discuss e.g.~the societal impact\footnote{We publish the code with the camera-ready version.}.

\section{Related work}

% In this section, we delve into related work and cover methods to internally and externally revise transformer models.
LLMs have been shown to possess the capabilities to perform basic reasoning and represent world knowledge \cite{petroni-etal-2019-language,rogers-etal-2020-primer,heinzerling-inui-2021-language}.
They have also been shown to contain a moral direction, i.e.~have human-like biases of what is right and wrong to do or, in other words, reflect some form of ethical and moral norms of society \cite{schramowski2022nmi_moral}.
% In this work, we want to utilize these basic capabilities and build our approach on top of this.
However, such models have flaws, e.g.~inconsistency in the generated representations or generally erroneous representations. Hence, a multitude of works has targeted revising incorrect model behavior, which can be subdivided based on how the revision is utilized.

\begin{figure*}[t!]
    \centering
    \includegraphics[width=0.7\textwidth]{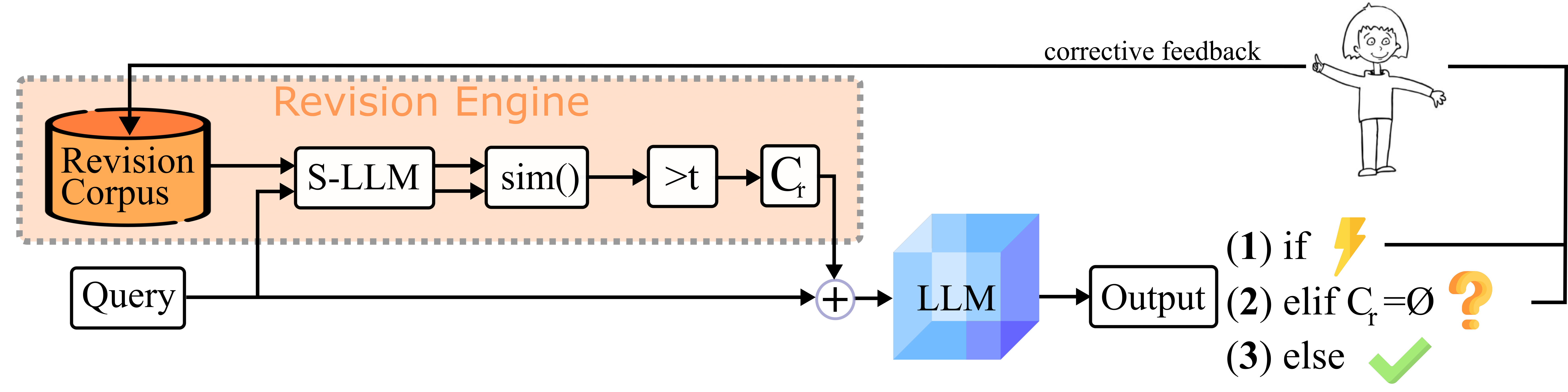}
    \caption{RiT architecture. The input to a base LLM is augmented with external context from the revision engine. The context is determined via similarity to the query in a sentence embedding space. If the model prediction is incorrect, i.e.~not aligned with the user's values, or the model is uncertain about the prediction, i.e.~no context was found, the user can interact with the model and provide corrective feedback.}
    \label{fig:architecture}
\end{figure*}

\paragraph{Internal Revision.}
One standard revision technique is to internally update the model parameters making them parametric approaches. Usually, the model parameters are \textit{fine-tuned} on new data \cite{howard-ruder-2018-finetuning}, while there are also approaches that completely train a model from scratch \cite{forbes-etal-2020-social}.
\cite{de-cao-etal-2021-editing} and \cite{delphi_jiang} have shown that fine-tuning a model's parameters on new corrective data helps revise knowledge. Nevertheless, (re)training a large-scale model is very costly, making it infeasible in a continual learning setting and lacking the capabilities for individual customization. 
Three promising and parameter-efficient approaches that reduce the revision cost are \textit{adapter tuning} \cite{houlsby_adapter}, \textit{bias tuning} \cite{ben-zaken-etal-2022-bitfit}, and \textit{prompt tuning} \cite{lester-etal-2021-prompttune}. Especially for prompt tuning, there is yet no single common taxonomy, and the latest research proposes several variants \cite{liu-etal-2022-p}.
% , such as hard, soft, p- and prefix tuning \cite{liu-etal-2022-p}. 
Although these methods are more parameter-efficient, being parametric means they are learning-based approaches which overall still require a large dataset for training as well as a careful hyperparameter search.
In contrast, \cite{meng2022locating} try to find the location of factual knowledge in the model, i.e.~the location of neurons that activate and attribute most to a given fact. The located neurons can be manipulated to conduct \textit{knowledge editing}. However, this and similar methods \cite{de-cao-etal-2021-editing} so far only work for factual knowledge which the model has already seen during training. 

\paragraph{External Revision.}
In our work, %, we want to empower individualistic model revision and hence focus external techniques. More precisely, 
we wish to make use of the latest results in prompt tuning without learning parameters and instead employ active prompt designing. 

As a first step, there are encouraging results in in-context learning \cite{garg_incontext,petroni2020how} indicating that adding context to the prompt can help a model learn from the given context and thus influence the inference step to a query. This can be described as \textit{prompt designing}. To avoid a tedious manual prompt designing, we make use of information retrieval in order to automate the contextualization of the query with relevant information. Thereby, revising the model behavior is non-parametric. The model architecture is augmented with a knowledge base in which the information is stored. Given an input query, this approach retrieves relevant information from the knowledge base and provides context to the query. The contextualization can occur directly in the input \cite{rag_lewis} or later in hidden layers through cross-attention \cite{retro_borgeaud}. However, previous work on in-context learning with information retrieval relies on (parametrically) tuning the retrieval. In contrast, we avoid (end-to-end) training the retrieval. Furthermore, we go beyond factual knowledge revision and do not rely on large (factual) data corpora like Wikipedia. In contrast to these, other works focus on moral revision through in-context learning. Madaan~\etal~\cite{madaan-etal-2022-memory} investigate teaching a model moral values, however they assume an indefinite memory size and context length where we specifically focus on sparsity in the amount of user feedback as well as context length. Both of which depict more realistic settings. Similarly, Jin~\etal~\cite{jin2022ruleBased} try to extract general rules about morality and when to break those. This approach, however, does not allow for human-in-the-loop revision and particularly lacks the possibility of flexible personalization to the individual which is a vital component for real-world deployment.

\paragraph{Interactive Revision.}
Rather, we propose to make use of user interactions, in which a user actively controls and revises the information stored in the revision engine. In many real-world use cases, there is only little data available, or the data is subjective, i.e.~there is no single overall true statement. Hence, there is ongoing work in interactive learning to revise a model through user interactions \cite{google_sparrow_22}. Even beyond, the XIL framework \cite{Teso2019_CAIPI,Schramowski20nmixil,xiltypo_friedrich} bases model revision by users not only on the prediction but also on the explanation. \cite{instructgpt_ouyang} also propose to revise a model through iterative user interaction with fine-tuning, i.e.~to rely on multiple parametric learning steps.
Our approach is also similar to (interactive) \textit{case-based reasoning} approaches \cite{prototrex_friedrich}, as RiT uses a similar case to augment the prompt and thereby guide the model output and a user can interact on the similar cases provided.
% Here, the user interacts with the model and controls the information stored in the knowledge base.
RiT builds on user interactions to improve model revision specifically in data regimes with sparsity and subjectivity.
% We aim to combine interaction with information retrieval to revise a model according to user preferences.

\section{Method}
% In the following, we describe RiT's architecture and properties.

\paragraph{General Architecture.}
In Fig.~\ref{fig:architecture}, we present the Revision Transformer (RiT) architecture. LLMs are commonly pre-trained on huge amounts of (uncurated) data. In order to efficiently update a model without costly parameter retraining, we propose to use an external revision engine. 
% The main components are the revision engine and the LLM. 
The general pipeline has the following steps. First, the query is passed to the revision engine. In order to find relevant information in the revision engine, we map query $q$ and entries $r_i$ from the revision corpus $\mathcal{R}$ into an embedding space, here with a sentence-level LLM (S-LLM): $e_x = \text{S-LLM}(x)$. Next, we measuring the similarity between their embedded vectors. Then we choose only the nearest-neighboring contexts, $C_r$, to be prepended to the query if they exceed a similarity threshold $t$, yielding
\begin{equation}
    C_r = \{r_i \in \mathcal{R} \; | \; \text{sim}(e_{r_i},e_q)>t\}
\end{equation}
Finally, this augmented prompt is fed into the LLM, which in turn generates an answer. Depending on the output, a user can give feedback to the model, i.e.~by removing, adding, or updating information in the revision engine. Overall, the revision engine can be integrated easily into any LLM off the cuff without training any parameters. We provide a pseudo code in Alg.~\ref{alg:rit}.
% \begin{align}
%     C_r &= \{r_i \in R if sim(r_i,e_i)>t\} \\
%     C_r &= argmax_{C'\subset C_r,|C'|=c}\sum\nolimits_{r\in C'}r
% \end{align}

\paragraph{Revision Engine.}
As previously described, the relevant contexts, $C_r$, are determined by detecting the nearest neighbors of the query within the revision corpus. 
%Furthermore, the similarity metric we use in the sentence embedding space is the cosine similarity.
In this course, we choose a threshold $t$ to retrieve only passages with a minimum similarity, i.e. with high relevance.
% regarding the query so that they contain relevant information for the revision. 
Next, the number of added contexts, i.e.~the size of $C_r$, can be controlled with $c$. It describes how many neighboring contexts will be considered as prefix and the most similar ones are selected:
\begin{equation}
    C_r = \text{argmax}_{C_r'\subset C_r,|C_r'|=c}\sum\nolimits_{r\in C_r'}r
\end{equation}
Both $t$ and $c$ are hyperparameters and require a careful selection. 
% Fortunately, LLMs themselves are able to identify context relevance \cite{petroni2020how}, discussed later.

% \begin{table}[t]
%     \centering
% 		\begin{tabular}{c|l}
%         augmented & \{Context\} Question: \{\textit{Query}\} Answer: \\ 
%         \hline
%         plain & Context: \{\textit{Related Context}\} \\ 
%         qna & Question: \{\textit{Related Context}\} Answer: \{\textit{Related Answer}\} \\ 

%         % \multirow{2}{1pt}{\textbf{(1)}}&
%         % \colorbox{cyan!30}{The food}was delicious and the service great\\
%         % &$\;$The\colorbox{cyan!30}{food was}delicious and the service great\\\hline
%         % %\parbox[t]{2mm}{\multirow{2}{*}{\rotatebox[origin=c]{90}{Attn.}}} 
%         % \multirow{2}{1pt}{\textbf{(2)}}&
%         % \colorbox{cyan!30}{The}food\colorbox{cyan!30}{\makebox(13,6){\vspace{-1.5pt}was}}delicious and the service great \\
%         % &$\;$The\colorbox{cyan!30}{food}was\colorbox{cyan!30}{delicious}and the service great
%     \end{tabular}
%     \caption{Prompt design with different prompt augmentation techniques}
%     \label{tab:contextualization}
% \end{table}

\paragraph{Contextualization.}
The generated output of an LLM depends on the way it is trained and on how the input prompt is designed.
% Tab.~\ref{tab:contextualization} depicts our prompt augmentation technique. 
We choose the general prompt design to be ``\texttt{Question}: \{query\} \texttt{Answer}:''. Supposing an input needs revision a RiT chooses the nearest neighbor in the revision engine and prepends it to the prompt, i.e.~``\{context\} \texttt{Question}: \{query\} \texttt{Answer}:''. In that case, there are two ways to integrate the context, either (i) with ``\texttt{Question}: \{context\} \texttt{Answer}: \{context answer\}'' or (ii) with ``\texttt{Context}: \{context\}''. To enable a more fine-grained contextualization, it is also possible to prepend multiple nearest neighbors to steer the generation process in the desired direction. 

Let us illustrate the contextualization by example. The question ``Should I travel by plane?'' turns into ``\texttt{Question}: Should I travel by plane? \texttt{Answer}:'' via our prompt designing. Assuming the nearest neighbor is ``Traveling by plane is bad for the environment.'', the contextualized prompt is ``\texttt{Context}: Traveling by plane is bad for the environment. \texttt{Question}: Should I travel by plane? \texttt{Answer}:''.

\begin{algorithm}[t]
\caption{The algorithm takes as input $q$, $\mathcal{R}$, $t$, and $c$.}
\begin{algorithmic}[1]
% \REQUIRE 
\STATE $C_r=\emptyset$
\STATE $e_q \gets \text{S-LLM}(q)$
\FOR{$r \in |\mathcal{R}|$}
\STATE $e_r \gets \text{S-LLM}(r)$
\IF {sim$(e_{r},e_q)>t$} 
  \STATE \texttt{ADD} $r$ to $C_r$
\ENDIF 
\ENDFOR
\IF{$|C_r|> c$} 
  \STATE \texttt{PICK} top $c$ elements of $C_r$, with argmax$(\text{sim}(e_{r},e_q))$
\ELSIF{$C_r=\emptyset$}
    \STATE \texttt{ASK} for user feedback and \texttt{ADD} to $ \mathcal{R}$
\ENDIF 
\STATE $q \gets C_r + q$ \COMMENT{prepend $C_r$ to $q$}
\STATE $\text{output} \gets \text{LLM}(q)$
\IF{output incorrect} 
  \STATE \texttt{ASK} for user feedback
\ENDIF 
\end{algorithmic}
\label{alg:rit}
\end{algorithm}

\paragraph{Classification with RiTs.}
In general, RiTs employ a sequence-to-sequence (seq2seq) transformer model and are therefore suited for text generation. However, RiTs can also be applied to classification tasks, requiring a slight adjustment. In standard classification tasks, the number of possible predictions is predetermined, while the output size of RiTs covers the vocabulary size times the number of generated tokens. Hence, a text-to-class mapping is required to approximate the predicted class from the generated text. 

% In general, applying RiTs to moral norms can be realized with binary classification, in which one class represents good/ moral and the other bad/ immoral. However, a RiT employs a sequence-to-sequence (seq2seq) model which generates language, so the number of possible predictions is actually the vocabulary size times the number of generated tokens. Hence, \ff{we} apply Jiang et al.'s text-to-class mapping\footnote{\url{https://github.com/liweijiang/delphi}} to estimate the binary polarity of the generated output. %\url{https://github.com/liweijiang/delphi/blob/master/script/evaluate/text2class.py}}.

Furthermore, a RiT contains an additional output state besides the generated tokens. If it finds no (relevant) context, \cf Fig.~\ref{fig:architecture}, it lets the user know and thereby exhibit its uncertainty, which we describe in more detail in the next paragraph.
% This puts the user (again) in the loop. They can now provide further feedback to the model by filling the revision engine such that human and machine make a step to cooperation \cite{Dafoe2021nature}.
\begin{table*}[t]
    \begin{center}
    \caption{RiT outperforms its baseline model through (iterative) user revision. NLG scores on the CNB test set.  %Sparseness investigation of user interactions with RiT, showing that fewer interactions suffice but more high-quality interactions still help improve. 
    RiT improves alignment of moral norms building on a T0 baseline with a large revision corpus (second row). A first iteration of user interaction results in a subset (7.5\%) of the train set ($i$RiT, third row) which already suffices to revise the model. Another iteration, i.e.~a few more user interactions, on non-contextualized examples ($i^2$RiT, fourth row) improve the model performance further. Best (``$\bullet$'') and runner-up (``$\circ$'') values bold.}
    \centering
        \centering
        {\def\arraystretch{1.1}
        \begin{tabular}{r|ccccccc}
             & \#feedback ($\downarrow$) &  Bleu-1 ($\uparrow$) & Bleu-3 ($\uparrow$) & Rouge-L ($\uparrow$) & METEOR ($\uparrow$) & Bertscore ($\uparrow$) & Acc. ($\uparrow$) \\
             \hline
            T0 & -- &  0.46 & 0.25 & 0.56 & 0.33 & 0.64 & 0.63 \\
            $\text{RiT}$ & $398\,468$ & $\circ$\textbf{0.77} & $\circ$\textbf{0.65} & $\circ$\textbf{0.79} & $\circ$\textbf{0.68} & $\circ$\textbf{0.87} & $\circ$\textbf{0.86} \\
            $i\text{RiT}$ & $\bullet$\textbf{29\,825} & 0.76 & 0.63 & 0.78 & 0.67 & 0.86 & $\circ$\textbf{0.86} \\
            $i^2\text{RiT}$ & $\circ$\textbf{32\,912} & $\bullet$\textbf{0.80} & $\bullet$\textbf{0.69} & $\bullet$\textbf{0.82} & $\bullet$\textbf{0.72} & $\bullet$\textbf{0.90} & $\bullet$\textbf{0.91}
        \end{tabular}
        }
        \label{tab:results}
    \end{center}

    \end{table*}
    
\paragraph{Interaction Protocol.}
There are three distinct situations that different model predictions can lead to (\cf Fig.~\ref{fig:architecture}). First (1), the predicted class is incorrect, i.e.~misaligned with the user's view (\imgsmall{figures/intro_lightning.png}). In this case, the model requires corrective user feedback, i.e.~by expanding or updating the revision corpus. In the second case (2), the model provides a prediction that might be correct or not; importantly, however, no context was found, exhibiting that the model is uncertain about its prediction (\imgsmall{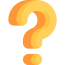}). This encourages the user to interact further with the model by providing missing context to the revision corpus. In the last case (3), relevant context is found in the model's revision corpus, and the prediction is well aligned with the user values (\imgsmall{figures/intro_check.png}).

\section{Experiments}
% In the following, we provide our experimental evaluations of RiT.

% \begin{table}[t]
%     \centering
%     \small
% 	{\def\arraystretch{1}\tabcolsep=6.pt
%     \begin{tabular}{l|cccccc}
%          &  Bleu-1 ($\uparrow$) & Bleu-3 ($\uparrow$) & Rouge-L ($\uparrow$) & METEOR ($\uparrow$) & Acc. ($\uparrow$) & \#feedback\\
%          \hline
%         T0 &  0.46 & 0.25 & 0.56 & 0.33 & 0.63 & -- \\
%         $\text{RiT}_\text{T0}$ & \textbf{0.77} & \textbf{0.65} & \textbf{0.79} & \textbf{0.68} & \textbf{0.86} & 398\,468
%     \end{tabular}
%     }
%     \caption{NLG scores on the test set from the commonsense norm bank dataset. Higher is better; best values in bold. RiT models outperform the baseline model through user revision.}
%     \label{tab:results}
% \end{table}

% \subsection{Experimental Protocol}
\paragraph{Model and Data.}
We conduct our experiments on a T0 model \cite{sanh2022T0}, a variant of T5, i.e.~an LLM, that is zero-shot able which in turn facilitates in-context learning.
In our experiments, we evaluate the auto-regressive generation of language from LLMs. For this seq2seq generation, we use sampling with top-k and set $k \!=\! 5025$, equal to $10\%$ of the vocabulary, and set the temperature to $0.1$. For finding relevant queries in our revision engine, we employ an S-LLM\footnote{\url{https://huggingface.co/sentence-transformers/sentence-t5-xl}} that is based on the same variant of T5. Furthermore, we use cosine similarity as a similarity measure. If not stated otherwise, we set $t\!=\!0.875$ and $c\!=\!1$. We use variant (i) for contextualization.
We apply RiT to the Commonsense Norm Bank (CNB) \cite{delphi_jiang}, which consists of multiple previously released datasets about morality. Here we focus on the oracle-like "agreement" section of the dataset.
% as it suits best. 

\paragraph{Feedback.}
Our approach builds on user feedback which is often only limited. For this purpose, we simulate user feedback in this work. Usually, datasets are provided with different splits, one for training, one for validating, and one for testing. Since RiT is a non-parametric approach, the training and validation sets are not required for learning. Instead, we can use both datasets to simulate user feedback and the test set for evaluation purposes.

\paragraph{Evaluation of Generated Answers.}
Evaluating the quality of natural language generation (NLG) is challenging and a research area in itself. One challenge is that there exists no single best metric, and a plethora is provided by current research, each with its individual pros and cons. We, therefore, utilize a set of metrics in order to broadly evaluate a model's generated answers. 
First, we use standard NLG scores \cite{sai2022Survey} like BLEU, ROUGE, and METEOR, which are n-gram based. Secondly, we investigate the cosine similarity in a sentence embedding space\cite{zhang2020bertscore}. Lastly, we apply a task-specific metric in the spirit of \cite{delphi_jiang}. Specifically, we calculate the binary polarity accuracy score. 
To do so, we apply Jiang et al.'s text-to-class mapping\footnote{\url{https://github.com/liweijiang/delphi}} to approximate the polarity of the generated text.
% the declarative (``yes''/ ``no'') and 
% the justification (``it is (not) okay'') part. 
% Two labels have matching polarity iff both parts are aligned in terms of polarity. 
% In order to estimate the polarity of the justification part, we use the above-mentioned text-to-class mapping.

\paragraph{Setup.}
We use the CNB dataset and show that the basic LM can only unsatisfactorily answer these moral questions. In order to teach the model, we keep the LM fixed and simply add an external revision engine. For our experiments, we have three different data setups and evaluate all on the test set (1) In our first approach, we simulate unlimited amount of user feedback and use the whole train data to fill the revision corpus. (2) In the second approach, we investigate sparsity in the amount of user feedback. Therefore, the revision corpus is empty and gradually filled with user feedback, which depicts the first ``real'' interactive iteration. We simulate the user feedback again with training examples that classify the (so far untouched) validation set best. (3) Lastly, we apply a second interactive iteration. Here, we use the data from (2) for the revision corpus and extend it with further feedback from the validation set. The remaining validation examples that could not be classified correctly with the training data are added to the revision corpus. 
For the classification, we employ the polarity accuracy as it is task-specific and models human preferences best.

%\subsection{Experiments}
\subsection{Aligning Moral Norms via \textit{RiT}}
% \paragraph{Revising LLMs}
\paragraph{Revising LLMs.}
In the initial experiment, we illustrate the facets of RiT and depict the demand for model revision.
To start with, we describe a basic use case; here, an LLM is used as an oracle to answer questions about morality. As a baseline we use a T0 model without a revision engine. At the same time, we use our RiT model, which is based on the same T0 model but on top of that utilizes the revision engine. 
We compare their performance on the CNB dataset. 

The top row of Tab.~\ref{tab:results} shows the performance of the baseline model. One can clearly observe that the default T0 model is not well aligned with the user values represented in the dataset. The generated answers only align in roughly 60\% of the examples with moral norms in terms of the accuracy metric. Nevertheless, for a baseline, this is a noteworthy performance as this LLM was never explicitly trained for this task and confirms previous findings about an LLM's general ability to contain a moral direction \cite{schramowski2022nmi_moral}. 

\begin{table*}[t]
    \begin{center}
    \caption{RiT evaluation (NLG scores on CNB test set), showing that RiT can be applied to various models (e.g. Bloom). Best values bold.}
    \centering
    {\def\arraystretch{1.1}
    \begin{tabular}{r|ccccccc}
     & \#feedback &  Bleu-1 ($\uparrow$) & Bleu-3 ($\uparrow$) & Rouge-L ($\uparrow$) & METEOR ($\uparrow$) & Bertscore ($\uparrow$) & Acc. ($\uparrow$) \\
     \hline
    Bloom3b & -- & 0.27 & 0.02 & 0.30 & 0.27 & 0.66 & 0.67 \\
    $\text{RiT}_\text{Bloom3b}$ & $398\,468$ & \textbf{0.50} & \textbf{0.22} & \textbf{0.54} & \textbf{0.44} & \textbf{0.76} & \textbf{0.82} \\
    \cdashline{1-8}
    Bloom176b & -- & 0.43 & 0.08 & 0.50 & 0.34 & 0.68 & 0.71 \\
    $\text{RiT}_\text{Bloom176b}$ & $398\,468$ & \textbf{0.56} & \textbf{0.26} & \textbf{0.60} & \textbf{0.52} & \textbf{0.86} & \textbf{0.90} \\
    \end{tabular}
    }
    \label{tab:results_bloom}
    \end{center}
\end{table*}

However, although the general moral direction may be given, this is still a serious alignment gap. We desire a model to be able to align to a high degree with user values without tediously tuning the parameters. In order to address this gap, we employ RiT. That means we still utilize the same baseline LLM (T0) and extend it with the, so far untouched, training data to fill the revision engine\footnote{A baseline model can be viewed as a RiT too with an empty revision engine.}. As can be seen in the second row of Tab.~\ref{tab:results}, with the help of the revision engine and without training any parameters, our RiT model improves accuracy off the cuff by more than 20\% compared to the baseline and more than doubles several of the NLG scores.  

This result can be found with other components as well. Tab.~\ref{tab:results_bloom} shows that RiT works well will other LLMs\footnote{The same is true for other S-LLMs.} (open-source Bloom \cite{Muennighoff2022CrosslingualGT}), too. Both RiT variants beat their respective baseline by a large margin. Notably, $\text{RiT}_{T0}$ and $\text{RiT}_\text{Bloom3b}$ outperform the Bloom176b baseline, although they have much fewer parameters. This emphasizes the utility of a revision engine. Comparing Bloom to T0, its superior performance can be explained by T0 being specifically optimized for zero-shot tasks, i.e.~in-context learning, in contrast to both Bloom-based RiT models. Still, they perform well on the task.

The shown findings on moral data coincide well with previous work on factual data \cite{petroni2020how}, and we conclude that RiT meets our basic expectations to improve model alignment with user values.
% , which presented that the augmentation of a prompt with context improves the task-specific performance .

% \paragraph{User Revision}
% \paragraph{RiT in Real-world Setting}
\paragraph{RiT vs Small Data.}
In the initial experiment, the revision engine comprised the full training data. However, such a large number of contexts is rarely available in real-world use cases. Furthermore, many of the datasets present in machine learning stem from Western cultures. What if a user wants to revise a model according to any other specific culture?

Thus, as subjective data and user interactions are usually scarce, we now consider the performance of RiTs in relatively small data regimes. In the previous experiment, we filled the revision engine with the full training dataset. Instead, we here simulate a more realistic scenario for collaborative user feedback by selecting certain training examples by means of the validation set and regard this as a first interactive iteration ($i$RiT). This way, the validation set acts as a proxy for the users' selection. We pick only those examples of the training set that help classify the validation set correctly and discard all others. As a measure of correctness, we choose the task-specific polarity accuracy. With this procedure, the revision engine size shrinks from 398\,468 to 29\,825, i.e.~down to 7.5\% of the original data size. Interestingly, as 
% Tab.~\ref{tab:simulation_results} 
the third row in Tab.~\ref{tab:results} shows, the performance is on par with the RiT model that had utilized the full train data. With this, we confirm the function of RiT in relatively small data regimes, showing that RiT is suited for more realistic use cases. % and cases that involve user interaction.

% \paragraph{Closer Inspection: Similarity \& Performance.}
\paragraph{Taking a Closer Look at RiT.}
The previous results are promising for RiT's ability to edit a model non-parametrically, particularly given scarcer data. When investigating the previous results further, one can observe that RiT performance is even better than at first sight. Specifically, on closer inspection, we find that $i$RiT performance is superior for examples where there is highly similar context available.

\begin{figure*}
    \centering
    \includegraphics[width=0.85\textwidth]{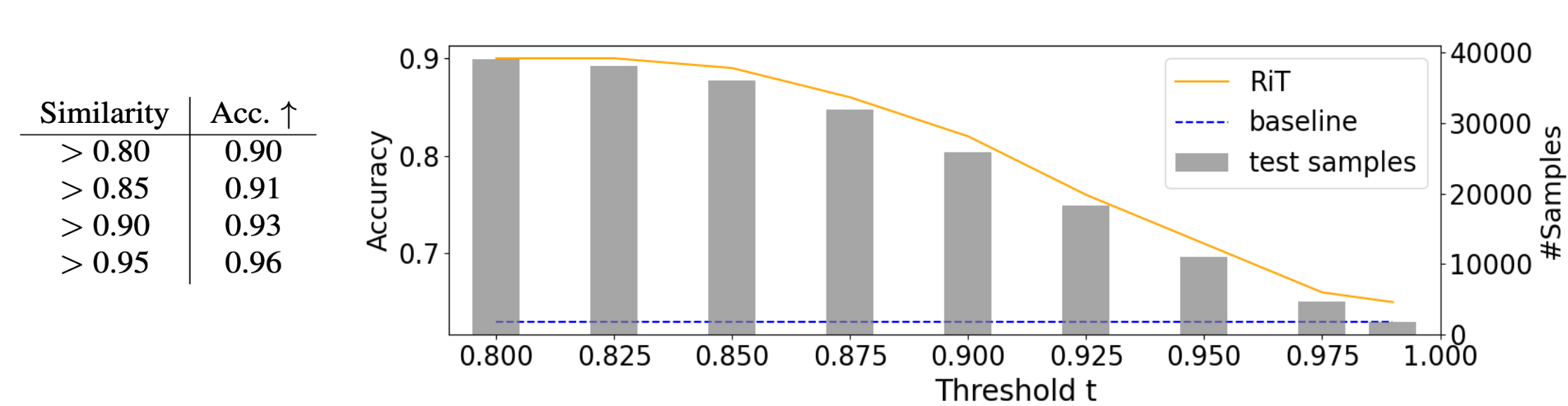}
    \caption{(left) The better the context, the better RiT. Examples revised with highly similar, i.e.~relevant, context align better with user values. (right) Examination of similarity threshold $t$. The graph depicts the number of contextualized test samples (gray), and the polarity accuracy for the baseline (blue-dotted) and RiT (orange). The higher the threshold, the fewer examples can be contextualized. Consequently, the RiT accuracy converges to the baseline performance with increasing $t$.}
    \label{fig:ablation_t}
\end{figure*}

%In other words, if we only inspect a subset of the test examples, namely the examples with contexts with a similarity of at least 0.8, 0.85, 0.9, or 0.95, we observe ...
Fig.~\ref{fig:ablation_t} (left) shows the results when we only inspect a subset of the test examples, namely those examples with context similarity of at least 0.8, 0.85, 0.9, or 0.95. We observe that the RiT performance increases for examples with increasing similarity. In other words: the higher the similarity between the retrieved context and the test example, the higher the model performance. 
% 
% Note that this performance is computed only on examples for which such a highly similar context is available. In fact, the number of relevant neighbors decreases with higher context similarity. This is due to that the revision engine so far does not solely contain perfectly aligned samples. As indicated by the results of the first experiment, the model achieves a performance of $86\%$ on all test examples, which is the result of the high performance for contextualized examples as well as the baseline performance for the non-contextualized examples. Ideally, we wish to arrive at a model with high performance for all examples through highly relevant context. 
% 
In summary, this evaluation provides evidence for the relationship between similarity and performance, showing that higher context similarity generally benefits RiT performance. Thus, one can expect the previous results to be even better had the context engine been provided more relevant contexts, where in the previous experiments, as a proxy, we had built the revision engine from the training data. Hence, in the following, evaluations we investigate additional means to ensure the retrieval of highly similar and relevant samples from the revision engine.
% Let us evaluate the practical implications of these findings in the following investigations and make a first step to address them. 

% \paragraph{A Naive Step Leads to Trade-off.} %ing off Performance and Relevance}
\paragraph{A Naive Step to Leverage Context Relevance.}
The detailed inspection of the model performance for contextualized examples leads us to how to ensure revision for all examples with only highly relevant, i.e.~similar, contexts. A naive step is to set a (high) similarity threshold, $t$, in the revision engine to receive only relevant context. 
% However, this comes at the expense of model accuracy, so a careful trade-off between context relevance and performance must be made. 

Fig.~\ref{fig:ablation_t} (right) describes the relationship between the similarity threshold $t$ and RiT's performance and the number of contextualized examples. The graph indicates that the number of contextualized examples decreases with increasing $t$. Also, RiT's performance decreases with increasing $t$, ultimately converging to the baseline performance at $t\!\approx\! 1$. This performance drop seems to contrast with previous similarity findings, showing that higher similarity yields higher performance. However, both results are in line as the performance is measured on different test (sub)sets. An increasing $t$ reduces the number of potential contexts for each test example as the revision corpus content, i.e.~the selected training samples, is diverse. In turn, only a portion of the test data benefits from RiT's contextualization.

More precisely, the overall model performance should be split into two parts: (i) relevant context is found (high performance), and (ii) no relevant context is found (baseline performance).
So, the overall model performance (orange) is the combined result of the contextualized (values in table) and non-contextualized (blue-dotted) examples.
% So, the overall model performance is the combined result of the high performance for contextualized examples and the baseline performance for the non-contextualized examples. 
With threshold $t$, we trade off the context relevance versus the number of available contexts. If $t$ is set too high, no neighbor can be retrieved anymore, yielding a RiT model that nearly behaves like its baseline (without revision engine), while setting $t$ too low can result in irrelevant context. Ideally, we wish to arrive at a model with high performance through highly relevant context for all examples. 

This experiment shows that a similarity threshold can be a naive step to only reinforce highly relevant contexts; however, this comes at the expense of finding adequate context.
\paragraph{User Interactions to the Rescue.}
As described before, a naive similarity threshold is limited by splitting the performance into two parts. However, a lack of relevant context ((ii) in the previous evaluations) is not necessarily at the expense of model performance, as we can go beyond a static threshold and address missing context with additional user interaction, i.e.~another interactive iteration ($i^2$RiT).
Moreover, even if the similarity threshold is set reasonably low, it is likely that some examples will not receive a context neighbor, particularly if the revision corpus is not filled with a large data amount. 

In the next evaluation, we, therefore, take a closer look at non-contextualized examples. To this end, we simulate a lack of context using a high threshold, resulting in many non-contextualized examples. At the same time, we employ a user's help for these examples by letting the user know that the model is uncertain, i.e.~no context was found. 
By that, the model uncertainty encourages the user to interact further and provide more relevant information to the revision engine, ultimately resulting in $i\!=\!2$ iterations of context extension. % Instead of similarity being at the expense of finding context, we propose to integrate the user into the pipeline once again to address a lack of context.
Specifically, for this task, we simulate user feedback again with the help of the validation set. The test data is evaluated with $i$RiT, i.e.~the subsampled revision engine and a threshold of 0.875. As a result, for roughly 5000 of the 40000 test examples, no relevant context is found. We examine these non-contextualized examples as the model expresses its uncertainty to the user. 
% treat those as cases in which the model expresses its uncertainty to the user. 
With the help of the validation set, we next simulate corrective user interaction and find relevant context for the non-contextualized examples in the validation set. This way, we find context for another 3000 examples and improve the RiT accuracy from 86\% to 91\%. At the same time, the revision engine is improved through an update, i.e.~expansion, of 3000 new examples yielding $i^2$RiT (\cf Tab.~\ref{tab:results}).
% With this setup, we can further improve the overall model performance achieving high alignment with user values. 

In conclusion, a model that is able to exhibit its uncertainty offers the option to address a lack of context, e.g.~through a similarity threshold, by iteratively incorporating the user into the revision pipeline. Hence, RiT presents an approach to conduct model revision and alignment beyond purely large-scale data-driven approaches.
% through the iterative integration of users.

\section{Discussion}
% Let us further discuss the implications of our approach and experimental findings.

\paragraph{What to Store Where?}
With this work, we want to illustrate a pathway for future AI models. Initially, we posed the question about what to store where and showed that non-parametric external model revision, e.g.~with RiT, addresses misalignment with user values.
However, our approach builds on large-scale pre-trained models, and they act as a lower bound, as the baseline performance pointed out. Our goal is to change paradigms from storing all information in the model parameters to also using external information. RiT is an enhancement of current LLMs, not a contradiction. In some cases, however, a value should not be overwritable through external interaction (e.g.~general Moral Code \cite{jin2022ruleBased}), and revising should require an organized parametric model update. Therefore, we propose the consideration of both external and internal revision modules in the model design choice. On a side note, they should be considered for other AI models as well as the RiT framework is not limited to transformers.

\paragraph{Societal Impact and Misuse.}
From an accessibility perspective, the sheer costs of retraining an LLM make it infeasible for nearly everyone except a limited number of companies or institutions. Hence, RiT offers a solution through an easily editable revision engine. In fact, replacing the whole engine is also merely \textit{Plug \& Play} such that nearly any user can insert their personal engine into RiT. This pushes boundaries back towards more democratic AI as users regain power that was recently ceded to a few tech companies. Yet, without an API available, end-users need certain skills as well to access LLMs. %on which we so far relied to provide trustworthy and good models. 

Moreover, recently ChatGPT has been banned in Italy by the government due to privacy concerns\footnote{available at \url{https://www.bbc.com/news/technology-65139406}}, i.e.~that it hurts the European General Data Protection Regulation\footnote{available at \url{https://gdpr-info.eu/}} (GDPR). This action calls for methods that enable a dynamical and flexible editing AI models. So, in order to be able to comply with the GDPR, RiT's revision engine offers again a first practical solution to this problem, as we showed. This is backed by other works on in-context learning \cite{deshpande2023toxicity,petroni2020how}, that personalization and overriding information through context memory is possible. And from the transparency perspective, a RiT can provide the retrieved context to the user (and the transformer attention weights), which, in turn, can increase model understanding.
% and interpretability.

On the other hand, what happens if a revision entry is inverted in its polarity and thus remains highly relevant? While human interaction has generally shown to be very helpful \cite{Ross2017_RRR,Teso2019_CAIPI}, at the same time, certain dangers can arise. For example, users with malicious intentions might target the model to misuse it. Hence, further detection mechanisms for malicious interaction are required. This is an active research area \cite{avoidtrolls_ju} that needs consideration when using human interaction. Nevertheless, an LLM's notion of what is right and wrong to do is still a first step to addressing (ir)relevant contexts. 

\begin{table*}[t]
    \centering
    \caption{Case study: effect of context relevance on generated prompts. Cosine similarity between context and query is given. If relevant context is retrieved, the model is revised. Even if not directly relevant context is retrieved, the model is able to identify relevance without threshold.}
    {\def\arraystretch{1.08}\tabcolsep=7.pt
    \begin{tabular}{ccl}
        %  prompt &  answer \\
         & Q & Do aircrafts pollute the environment? \\
         & A & No, they do not. \\
        \cdashline{2-3}
         0.94 & C+Q & Aircrafts pollute the environment. Do aircrafts pollute the environment? \\
         & A & Yes. They are a major source of air pollution. \\ 
        \toprule
         & Q & Are aircrafts bad for the environment? \\
         & A & No, they are good for the environment. \\
        \cdashline{2-3}
         0.90 & C+Q & Aircrafts pollute the environment. Are aircrafts bad for the environment? \\
         & A & Yes, they are bad for the environment. \\ 
        \toprule
        & Q & What is the fastest option to travel to New York? \\
        & A & The fastest option to go to New York is to take a direct flight. \\
        \cdashline{2-3}
        0.69 & C+Q & Aircrafts pollute the environment. What is the fastest option to travel to New York? \\
        & A & The fastest option is to fly. \\
    \end{tabular}
    }
    \label{tab:case_study}
\end{table*}

\paragraph{User Interaction.}
Throughout our work, we iteratively took advantage of user interactions. And yet, there are more options for incorporating a human user. Not only the model but also the user can express uncertainty. In such a case, a user might still have a notion of what might be relevant. The model, in turn, could provide the closest available context but below threshold $t$. A user can (refrain to) adopt this suggestion in order to facilitate filling in the missing context.

On the other hand, the cost of human labor requires consideration. User interaction is not free of effort and has limitations. For instance, if the threshold is set too high, the demand for user interaction increases as well. So, in general, a trade-off between model performance and human labor must be kept in mind. However, an important aspect of a RiT is its baseline performance of 63\%, such that only in the misclassified cases must the model ask for a revision, drastically reducing human labor. Furthermore, a RiT is especially powerful in cases where only a few revisions must be made. 

\paragraph{RiT and Misalignment.}
Furthermore, we want to discuss the weakness of each of the RiT components by looking at misclassifications. Let us consider its components individually: the revision corpus, the revision retrieval, and the LLM. In the experimental section, we addressed missing context in the revision corpus with user interaction.
Besides a lack of respective context in the engine, not finding relevant context can also be due to a suboptimal retrieval. If a neighbor is available but not selected, the context retrieval should be improved. 
Training the retrieval process optimizes the neighbor selection and helps find relevant context. 
In the last case, context is available and selected, but the LLM still generates an incorrect answer. This indicates that the LLM itself needs revision. These two deficiencies can be addressed with an update of the parameters, e.g.~utilizing one of the parameter-efficient techniques.
As previously mentioned, we regard the extension of RiT via such parametric revision techniques as a promising avenue.% for future work.

\paragraph{Threshold and Relevance.}
Here we wish to examine the context relevance further. Tab.~\ref{tab:case_study} shows in a qualitative case study that a threshold is not the only means to handle the relevance of contexts. In the first two cases, the context is relevant to the query, as indicated by the high similarity value. Moreover, the model is successfully revised through context. In contrast, the context in the third case is not directly relevant (similarity of ~0.7) and does not revise the model.
This suggests that even if the similarity threshold is set low, the LLM itself can be a means to identify and consider the relevance of the given context. If (accidentally) an irrelevant or not directly related context is provided, the model can still ignore the given context \cite{petroni2020how}. Actually, this relevance filtering can be found twice in RiT. RiTs employ the same basic transformer in the revision engine (S-LLM) and the language generation (LLM) which are both based on the same variant (T5). This way, RiTs possess a degree of inherent robustness for relevance. It is worth noting, that ignoring context in the wrong situation can also come along with certain downsides such as continuing to exhibit unwanted (biased) output.

\paragraph{Evaluating NLG.} In general, standard measures to evaluate NLG suffer from a semantic gap \cite{sai2022Survey}. For instance, our experiments uncovered that the LLM often generates the right justification but the wrong declarative part of the answer as a result of the \textit{negative question} problem. In other words, ``Shouldn't you do ...?'' can also be answered by ``No, you shouldn't'' while the actual ground truth is ``Yes, you shouldn't''. Humans often treat both answers as equivalent, a well-known finding for human communication \cite{negative_kamoen}, which is very certainly also represented in such a way in the large-scale pre-training data. As a means, we provide results on a set of various NLG scores to better and diversely evaluate the task at hand. To this end, we extended standard scores (Ngram-based \cite{kim-etal-2010-evaluating}) with task-specific (polarity accuracy \cite{delphi_jiang}) and semantic-focused (Bertscore \cite{zhang2020bertscore}) scores. So, while the absolute value of each score might be treated with a grain of salt, they still reflect helpful indicators for evaluating and the combination of many scores gives strong evidence for the function of our method. And in the real-world, humans generally decide what an adequate revision looks like and when to interact.

\section{Conclusion}
% Current transformer language models (LM) are large-scale models with billions of parameters. They have been shown to provide high performances on a variety of tasks but are also prone to shortcut learning and bias. Addressing such incorrect model behavior via parameter adjustments is very costly. This is particularly problematic for updating dynamic concepts, such as moral values, which vary culturally or interpersonally. In this work, we question the current common practice of storing all information in the model parameters and propose the Revision Transformer (RiT) employing information retrieval to facilitate easy model updating. The specific combination of a large-scale pre-trained LM that inherently but also diffusely encodes world knowledge with a clear-structured revision engine makes it possible to update the model's knowledge with little effort and the help of user interaction. We exemplify RiT on a moral dataset and simulate user feedback demonstrating strong performance in model revision even with small data. This way, users can easily design a model regarding their preferences, paving the way for more transparent and personalized AI models.
    
This work investigated the benefits of integrating a revision engine into transformer-based LLMs. We propose the Revision Transformer (RiT), question the current common practice of storing all information in the model parameters, and alternatively propose to extend current models with a revision engine. Our results indicate that this framework helps correct model behavior and align a model with user values. Moreover, RiTs iteratively employ user interaction to incorporate corrections with little effort achieving high value alignment.

While different languages often go hand in hand with cultural differences and differing moral norms, RiTs can also be employed in such subjective cases. In future applications, each language or cultural subgroup could e.g.~have its own revision engine built on top of a common LLM. Thus, an exciting pathway for future research is to evaluate RiTs for different cultures. An ultimate goal might be to set up a hub where each user can integrate their customized revision engine or collaborate with others to design models which are highly aligned with their values. Furthermore, with the rise of multilingual language models, it is also interesting to evaluate revision for different languages. As these models are trained on highly unbalanced data (English-focused), investigating whether revision works is crucial.

% improving information retrieval through a differentiable retrieving to enable end-to-end learning. Another exciting future work is applying RiT to go beyond moral norms. 

% Filtering the added revision entries, check whether it is malicious content, troll detection
% expand retriever entries with related entries through https://www.mdpi.com/2073-8994/14/8/1715 But shouldn't the model already know that related entries show use the same information?

\subsubsection*{Acknowledgments}
% We thank the anonymous reviewers. 
This work benefited from the ICT-48 Network of AI Research Excellence Center “TAILOR" (EU Horizon 2020, GA No 952215), the Hessian research priority program LOEWE within the project ``WhiteBox'', and the Hessian Ministry of Higher Education, Research and the Arts (HMWK) cluster projects ``The Adaptive Mind” and ``The Third Wave of AI'', and from the German Center for Artificial Intelligence (DFKI) project ``SAINT''.
%\clearpage

% \bibliographystyle{ecai.bst}
\bibliography{ecai.bib}

% \begin{algorithm}
% \caption{The algorithm takes as input query $q$, revision engine $\mathcal{R}$, languages models LLM and S-LLM, cosine similarity function sim, similarity threshold $t$ and the number of nearest neighbors $c$.}
% \begin{algorithmic}[1]
% % \REQUIRE 
% \STATE $C_r=\emptyset$
% \STATE $e_q \gets \text{S-LLM}(q)$
% \FOR{$r \in |\mathcal{R}|$}
% \STATE $e_r \gets \text{S-LLM}(r)$
% \IF {sim$(e_{r},e_q)>t$} 
%   \STATE \texttt{ADD} $r$ to $C_r$
% \ENDIF 
% \ENDFOR
% \IF{$|C_r|> c$} 
%   \STATE \texttt{PICK} top $c$ elements of $C_r$, with argmax$(\text{sim}(e_{r},e_q))$
% \ELSIF{$C_r=\emptyset$}
%     \STATE \texttt{ASK} for user feedback
% \ENDIF 
% \STATE $q \gets C_r + q$ \COMMENT{prepend $C_r$ to $q$}
% \STATE $\text{output} \gets \text{LLM}(q)$
% \IF{output incorrect} 
%   \STATE \texttt{ASK} for user feedback
% \ENDIF 
% \end{algorithmic}
% \label{alg:rit}
% \end{algorithm}

\end{document}